\documentclass[runningheads]{llncs}
\usepackage{graphicx}

\usepackage{tikz}
\usepackage{comment}
\usepackage{amsmath,amssymb} 
\usepackage{color}

\usepackage{times}
\usepackage{epsfig}
\usepackage{gensymb}

\makeatletter
\DeclareRobustCommand\onedot{\futurelet\@let@token\@onedot}
\def\@onedot{\ifx\@let@token.\else.\null\fi\xspace}

\def\eg{\emph{e.g}\onedot} 
\def\ie{\emph{i.e}\onedot} 
 
\def\etc{\emph{etc}\onedot}

\def\etal{\emph{et al}\onedot}
\makeatother

\newcommand{\Tref}[1]{Table~\ref{#1}}
\newcommand{\Eref}[1]{Eq.~(\ref{#1})}
\newcommand{\Fref}[1]{Figure~\ref{#1}}

\newcommand{\Cref}[1]{Chap.~\ref{#1}}
\newcommand{\Sref}[1]{Section~\ref{#1}}

\usepackage{colortbl}
\definecolor{steelblue}{rgb}{0.27, 0.51, 0.7}

\newcommand{\balpha}{\mbox{\boldmath $\alpha$}}

\newcommand{\be}{\begin{eqnarray}}
\newcommand{\ee}{\end{eqnarray}}
\newcommand{\bee}{\begin{eqnarray*}}
\newcommand{\eee}{\end{eqnarray*}}

\newcommand{\matrixb}{\left[ \begin{array}}
\newcommand{\matrixe}{\end{array} \right]}

\newcommand{\argmax}{\operatornamewithlimits{\arg \max}}

\newcommand{\app}{\raise.17ex\hbox{$\scriptstyle\sim$}}

\usepackage{enumitem}
\usepackage[normalem]{ulem}
\usepackage[ruled]{algorithm2e}
\usepackage{multirow}
\usepackage{array}
\usepackage{ctable} 
\usepackage{makecell}
\usepackage[capposition=bottom]{floatrow}
\usepackage{comment}
\DeclareMathAlphabet{\mathcal}{OMS}{cmsy}{m}{n}
\usepackage{textcomp}
\usepackage{dblfloatfix}
\usepackage{caption}
\usepackage{subcaption}
\usepackage{colortbl,xcolor}
\usepackage{booktabs}
\usepackage{mathtools}
\usepackage[flushmargin]{footmisc}

\setlist[itemize]{align=parleft,left=0pt}
\renewcommand{\paragraph}[1]{\vspace{1.5mm}\noindent\textbf{#1}\quad}

\definecolor{customgray}{rgb}{0.9, 0.9, 0.9}
\newcolumntype{g}{>{\columncolor{customgray}}c}
\newcolumntype{z}{>{\columncolor{customgray}}l}
\newcolumntype{?}[1]{!{\vrule width #1}}

\usepackage[accsupp]{axessibility}  

\usepackage[width=122mm,left=12mm,paperwidth=146mm,height=193mm,top=12mm,paperheight=217mm]{geometry}

\begin{document}
\pagestyle{headings}
\mainmatter
\def\ECCVSubNumber{100}  

\title{Audio-Visual Fusion Layers for Event Type Aware Video Recognition} 

\titlerunning{Audio-Visual Fusion Layers for Event-Aware Recognition} 
\authorrunning{A. Senocak and J.Kim et al.} 
\author{Arda Senocak*\inst{1} \and
Junsik Kim*\inst{2} \and
Tae-Hyun Oh\inst{3} \and
Hyeonggon Ryu\inst{1} \and
Dingzeyu Li\inst{4} \and
In So Kweon\inst{1}}
\institute{KAIST \and
Harvard University \and
POSTECH \and Adobe Research}

\maketitle

\begin{abstract}
Human brain is continuously inundated with the multisensory information and their complex interactions coming from the outside world at any given moment. Such information is automatically analyzed by binding or segregating in our brain. While this task might seem effortless for human brains, it is extremely challenging to build a machine that can perform similar tasks since complex interactions cannot be dealt with single type of integration but requires more sophisticated approaches. In this paper, we propose a new model to address the multisensory integration problem with
individual event-specific layers in a multi-task learning scheme. Unlike previous works where single type of fusion is used, we design event-specific layers to deal with different audio-visual relationship tasks, enabling different ways of audio-visual formation. 
Experimental results show that our event-specific layers can discover unique properties of the audio-visual relationships in the videos. Moreover, although our network is formulated with single labels, it can output additional true multi-labels to represent the given videos. 
We demonstrate that our proposed framework also exposes the modality bias of the video data category-wise and dataset-wise manner in popular benchmark datasets.
\end{abstract}

\vspace{-4mm}\section{Introduction}\label{sec:intro}
\begin{figure}[t]
\centering
        \includegraphics[width=.5\linewidth]{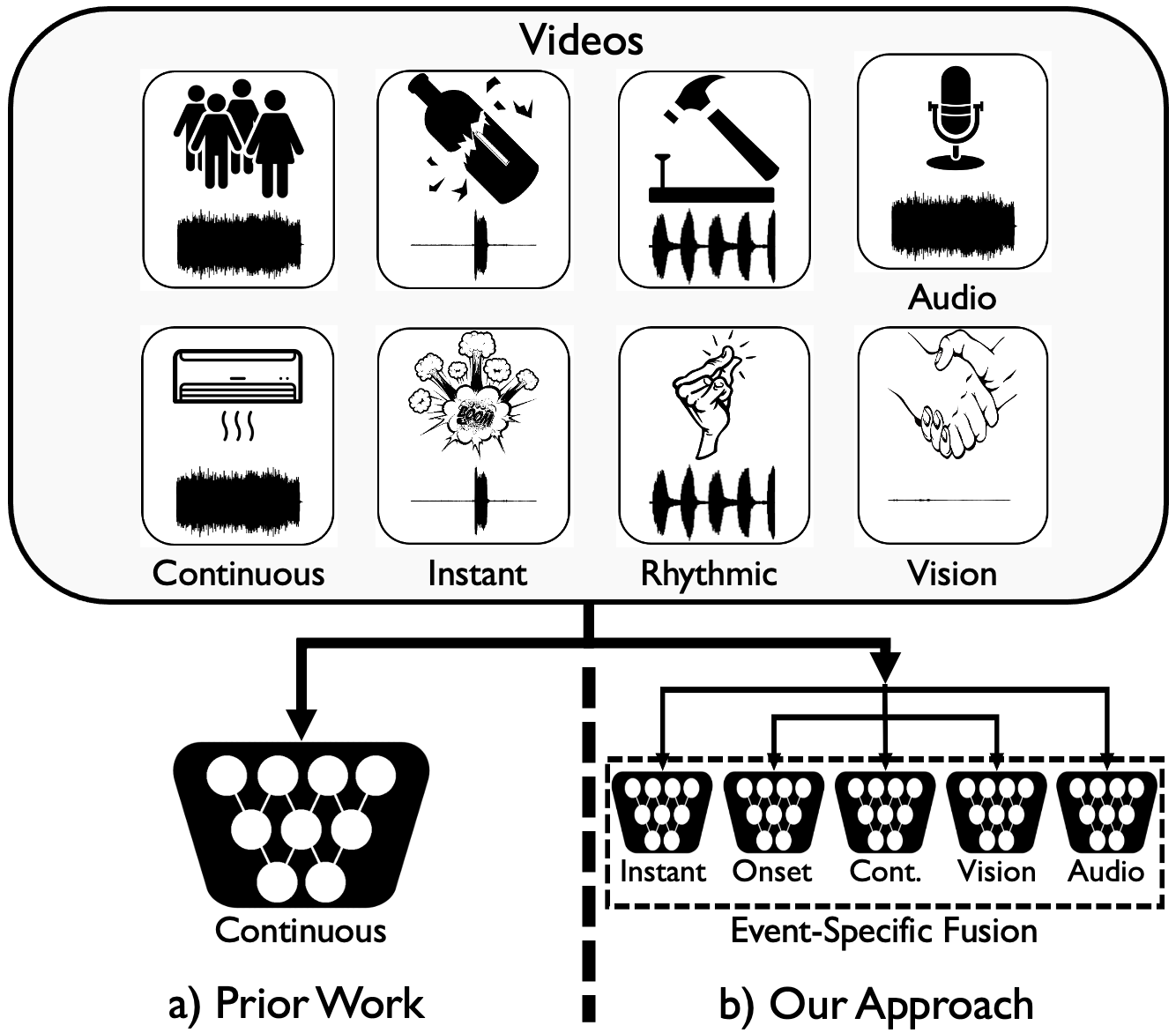}
\caption{\textbf{A conceptual difference between prior approaches and ours.} Multi-modal events can be based on various forms in in-the-wild videos; while some events might have continuous temporal correspondence between visual changes and accompanied audio, the others may have rhythmic, repetitive audio-visual events or a few isolated instant moments, \eg, a person snapping her fingers rhythmically with background music rhythm; the air conditioner is blowing continuously; or a volcano explodes in the footage.
Despite these complexities, prior approaches use one-size-fits-all fusion methods with barely considering diverse event-types. 
In contrast, we use multiple event-specific layers for better video understanding. 
\vspace{-4mm}}
\label{fig:second_teaser}
\end{figure}

Real-world events around us consist full of different multisensory signals and their complex interactions with each other. 
In-the-wild videos of real-life events and moments capture a rich set of multi-modalities and their complex interactions therein. 
Thus, it is essential to leverage multisensory information for better video understanding, but their diversity and complex nature make it challenging. While some of these events -- such as hearing an engine sound and seeing a race car in the video -- have corresponding multi-modalities, some others can have non-corresponding multi-modal signals. 
For example, sometimes sounds are produced by events that we do not see in the field of view (\eg, `voice over'), or some visual events do not make any distinctive sounds (\eg, `hand shaking'). 
Even though audio and vision signals are congruent, the way they relate are also different. All these events have different types of temporal regularity, as shown in~\Fref{fig:second_teaser}, which we call them as \emph{event types}.
That is, understanding video contents require to understand and properly deal with such diverse and complex associations and relationships.

In the computer vision field, there have been vast efforts, \eg ~\cite{korbar2018cooperative,owens2018multisensory,tian2018ave,wang2020multimodalHard,kazakos2019epicFusion,long2018pureAttention,xiao2020avSlowFast}, to implement a machine perception for multi-modal video analysis.
They model multi-modal fusion mechanisms with a fundamental assumption that all audio-visual events are highly correlated and continuous throughout the video. This uniform assumption is flawed in its ability to model real-world multisensory events as illustrated in~\Fref{fig:second_teaser}.\footnote{Let us consider the car video in~\Fref{fig:layers}. A group of spectators talk for a long time while recording the scene before race car passes, but the only useful moment for audio-visual integration is the short moment car passes by.}
The common fusion method, \eg, ~\cite{arandjelovic2017L3,zhang2021repetitive,korbar2018cooperative,wang2020multimodalHard}, which is concatenating or average-pooling both modalities over the entire sequence, smooths out sparse important moments and leads to wrong prediction (\eg, in a race event, `crowd' sound is dominant instead of `car sound' in terms of the global-pooling based fusion method). Thus, in contrast to the prior methods which all use an one-size-fits-all fusion, we present a new perspective that incorporates event-specific layers to improve video understanding.

We developed a model that incorporates multiple types of audio-visual relationships to improve video understanding.
Our development is motivated by the way humans perceive the world.
Humans are spontaneously capable of combining relevant heterogeneous signals from the same events or objects, or distinguishing a signal from one another if the source events of the signals are different; thereby, we are able to understand events around us.
Such multisensory integration has been widely studied as binding \emph{vs.} segregation of unimodal stimuli in cognitive science~\cite{kayser2015multisensory,spence2007multisensory,spence2011crossmodal}.
However, it is challenging to identify all possible types of binding and segregation for all existing event types in the world.
We postulate that most of existing events may be effectively spanned by combinations of a few dominant event types previously identified by the cognitive science studies, such as continuous, onset and instant event.

Motivated by these observations, we attach individual modality and audio-visual event-specific layers into the model in a multi-task learning scheme. Each layer is designed to look for different characteristics in the videos such as audio only, visual only, continuous, onset and instant event layers. We show that our method improves video classification performance and also enables reliable multi-label prediction. 
Our proposed model leads to better interpretability of the videos such as understanding the audio and visual signals independently or jointly based on the characteristics of the events as well as providing na\"ive modality confidence scores. This allows us to conduct interesting existing dataset analyses and potential applications such as multi-labeling, category-wise and dataset-wise modality bias, and sound localization.

\section{Related Work}

\paragraph{Cognitive Science.} Our work is motivated by the findings in the numerous biology, psychology, and cognitive science research investigations on multisensory integration and causal inference in the brain~\cite{su2014avRhythm,spence2007multisensory,spence2011crossmodal,kayser2015multisensory,nahorna2012bindUnbind,nahorna2015sceneAnalysis,spence2012book}.
Evidences in these studies show that brain solves two problems:
1) to bind or segregate the different sensory modalities depending on they originate from a common or separate causes (events or objects);
2) if they go together, how to integrate them properly. 
These studies suggest human brain uses different types of perceptual factors - such as temporal, spatial, semantic and structural - while integrating different sensory signals.
In our work, we take inspiration from these studies and formalize the multisensory binding \emph{vs.}~segregation and the causal inference problems by designing multisensory event-specific layers in multitask learning scheme.

\paragraph{Audio-Visual Representation Learning.} Recent years have witnessed significant progress in audio-visual learning and some used audio or visual information as supervisory signal to the other one~\cite{aytar2016soundNet,owens2016ambient,owens2016indicated} or leverage both of them in self-supervised learning to learn general representations assuming that there is a natural correspondence between them~\cite{arandjelovic2017L3,arandjelovic2018object,hu2019cluster,aytar2017see,korbar2018cooperative,owens2018multisensory}.
Self-supervised learning methods use different tasks such as correspondence~\cite{arandjelovic2017L3,arandjelovic2018object,aytar2017see}, synchronization~\cite{korbar2018cooperative,owens2018multisensory} or clustering~\cite{alwassel2020XDC}.
Furthermore, recent explorations use audio-visual multimodal signals as self-supervision to cluster or label the unlabelled videos~\cite{asano2020selavi,alwassel2020XDC}. These existing approaches assume that multisensory data is always semantically correlated and temporally aligned. As a result, they apply simple fusion techniques such as concatenation or average pooling.
However, in real world videos, multisensory data are not always naturally co-occurring.
Our work investigates more diverse multisensory relationships and propose different integration approaches in audio-visual events. 

\paragraph{Audio-Visual Activity Recognition.} Various deep learning approaches have been proposed to boost action recognition accuracy by incorporating audio as another modality~\cite{long2018pureAttention,korbar2019scsampler,kazakos2019epicFusion,xiao2020avSlowFast,gao2020listen2look,wang2020multimodalHard}. 
While most of the existing work simply concatenate the audio and visual features with naive approaches, Gao~\etal\cite{gao2020listen2look} uses multi-modal distillation from a video model to an image-audio model for action recognition. Besides these existing work, Wang~\etal\cite{wang2020multimodalHard} investigates that naive approaches may not be the optimal solution to training multimodal classification networks and proposes to use joint training by adding the two separate modalities with weighted blending. Our learning mechanism is similar to this approach in terms of multi-task joint training but our training scheme is applied to multiple event-specific layers to address proper multisensory integration. 

\paragraph{Broader Audio-Visual Learning Tasks.} Recent work on audio-visual learning use the natural correspondence between auditory and visual signals on different tasks than representation learning and action recognition, including audio-visual sound separation~\cite{ephrat2018looking,gan2020gestureSep,gao2018learning,gao2019coSep,gao2021visualVoice,afouras2020AVObjects,zhou2020sepStereo,zhao2019soundMotion,zhao2018soundPixels,xu2019minusPlus,tzinis2021audioScope}, sound source localization~\cite{senocak2018learning,arandjelovic2018object,senocak2019learningTPAMI,tsiami2020stavis,hu2020localization}, audio generation~\cite{zhou2018visual2sound,morgado2018spatialAudio,gao2019sound25D,zhou2019audioInpaint,yang2020left2right} and audio-visual event localization~\cite{lin2019seq2seq,tian2018ave,wu2019dualAttention}. Different from all the previous work, we focus on incorporating audio and visual modalities for multisensory integration without the assumption that they are always correspondent.

\vspace{-1mm}
\section{Approach}
\begin{figure}[t]
\centering
	\includegraphics[width=.8\linewidth]{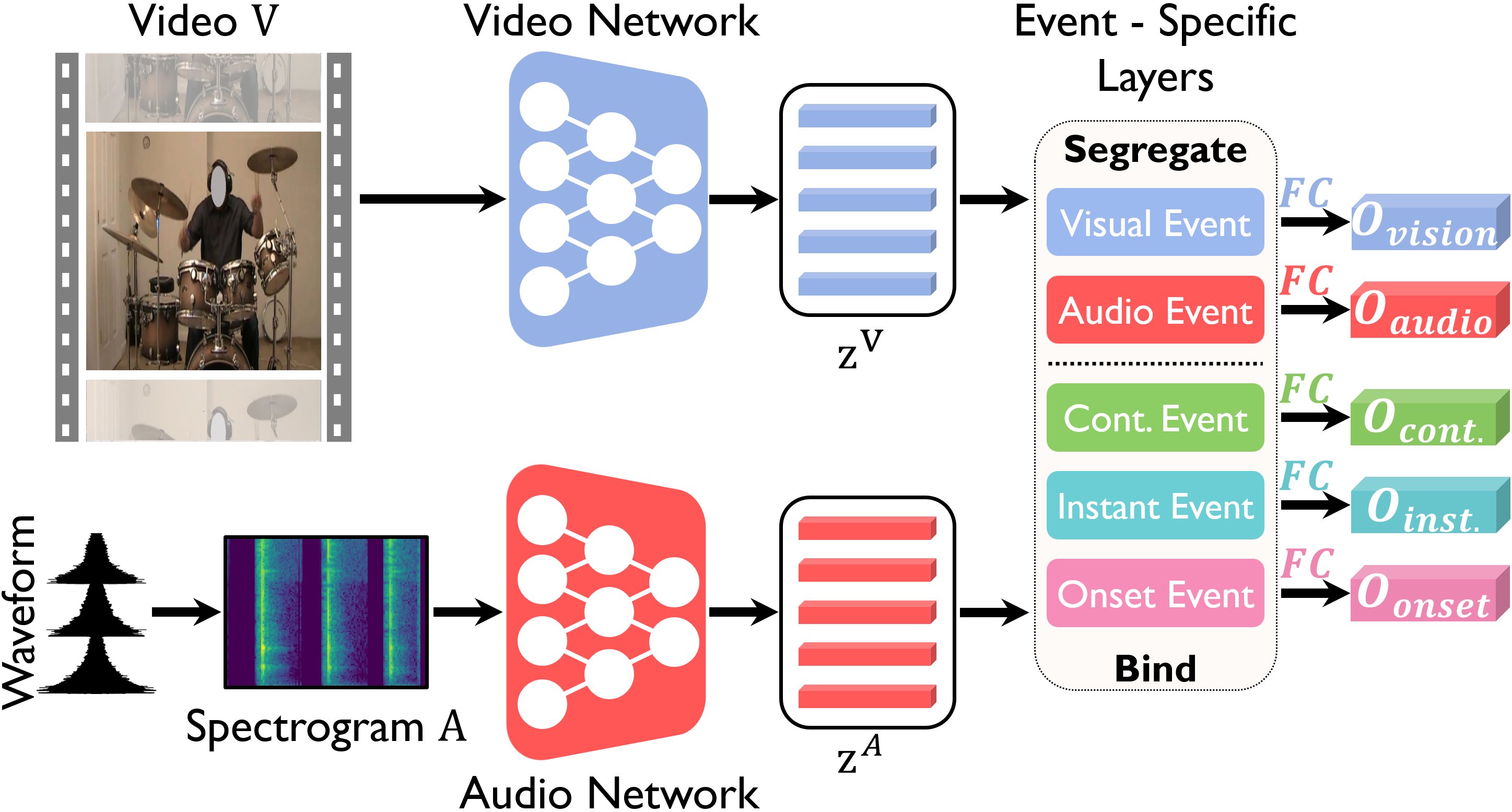}
	\caption{\textbf{Our multisensory framework.} The model consists of video and audio backbone networks that extract video-level features, $\mathbf{z}^\mathbf{V}$ and $\mathbf{z}^\mathbf{A}$, for each modality. The features are fed into \emph{Event-Specific Layers} for multisensory integration. Each layer processes the features individually and predicts a label category.\vspace{-2mm}}
\label{fig:pipeline}
\vspace{-2mm}
\end{figure}

\begin{figure*}[t]
	\begin{center}
		\includegraphics[width=1\linewidth]{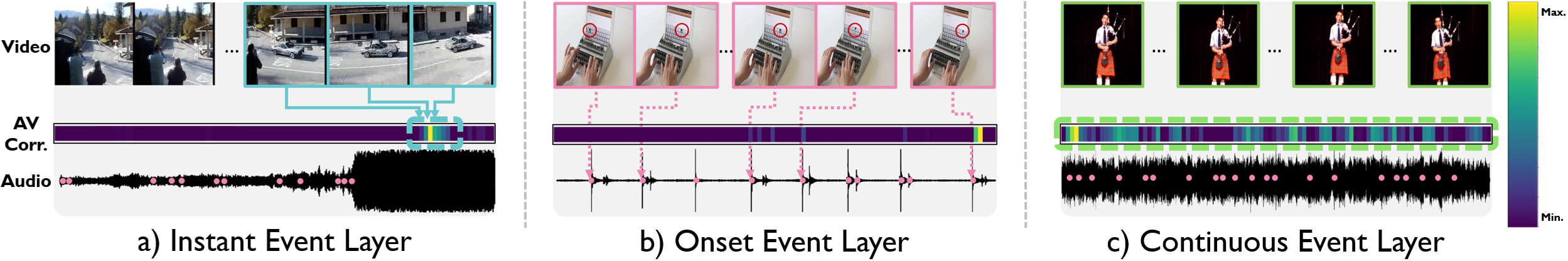}\vspace{-3mm}
	\end{center}
	\caption{{\bf Audio-Visual Event Layers.} We show how each audio-visual event-specific layers perform their pooling operation. 
	\emph{Instant Event Layer} picks moments where audio and visual features agree, highlighted by the AV Correspondence heatmap. 
	\emph{Onset Event Layer} only pools audio onsets (pink dots on the waveform) into the feature computation.
	\emph{Continuous Event Layer} adopts the traditional global average pooling with uniform sampling.\vspace{-2mm}
	}
	\label{fig:layers}
	\vspace{-2mm}
\end{figure*}

\subsection{Problem Formulation}\label{sec:problem_formulation}
The goal of our model is to understand and predict the accurate label that represents the video from the perspective of each multisensory layer. Most of the existing work~\cite{long2018pureAttention,korbar2019scsampler,kazakos2019epicFusion,xiao2020avSlowFast,wang2020multimodalHard} used a clip level classifier that took a short ($1$ or $2$ sec.) clip and then computed video-level predictions by averaging classification scores of each clip. These clip classifiers are learned by leveraging na\"ively fused (\eg, simple concatenation) audio-visual features with the assumption of audio and visual data are correlated and temporally aligned.

As aforementioned~(\Sref{sec:intro}), these existing approaches for video classification and understanding might be improved by considering more complex associations. First, audio and visual events in a video may not occur with a close association all the time. They can occur separately in each individual modality as well. Second, this audio-visual correspondences can have different characteristics such as continuous, rhythmic or isolated instant events~\cite{spence2007multisensory,su2014avRhythm}. Our proposed architecture addresses these concerns by using various multisensory layers.

\paragraph{Backbone Networks.} \label{sec:backbone} 
Given a video clip $\mathbf{V}$ with its corresponding audio $\mathbf{A}$, our backbone networks extract features for each modality. We use a two-stream architecture, that leverages each modality separately, similar to other existing audio-visual learning works. Our backbone networks take entire 10 sec. video and audio frames and extract features per-frame for each modality. The video network is a spatio-temporal network, similar to MC$x$~\cite{tran2018multimodalHard}. It takes a video $\mathbf{V}$ of $T$ frames as input and generates a video embedding $\mathbf{z}^\mathbf{V}$ with dimensions $T \times D$. The audio stream takes the log-mel spectrogram $\mathbf{A}$ of 10$T$ frames and passes it through a stack of 2D convolution layers to extract an audio embedding $\mathbf{z}^\mathbf{A}$ with dimensions $T \times D$ similar to video features. Thus, there is a corresponding audio feature for every video feature and we do not need any replication or tile operations to match audio and video feature dimensions.

\subsection{Multisensory Event-Specific Layers}\label{sec:task_layers}
To deal with different multi-modal event types, we design those expert layers as multi-task heads of the multi-modal network as shown in~\Fref{fig:pipeline}. Defining $i$ the index of each layer, the layer takes $\mathbf{z}^\mathbf{V}$ and $\mathbf{z}^\mathbf{A}$ from backbone networks and outputs video-level prediction ${O}_i$ by applying assigned task to itself. We explain each layer in detail below.

\paragraph{Continuous Event Layer.} This layer is designed to integrate audio and visual signals by performing temporal aggregation to each frame features from both modalities with the assumption of audio and visual signals are temporally correlated and aligned throughout the video. This temporal congruency between audio and visual signals play a key role for audio-visual sensory integration not only in cognitive science~\cite{spence2007multisensory,spence2011crossmodal} but also in audio-visual learning works~\cite{owens2018multisensory,tian2018ave,korbar2018cooperative,halperin2019dynamicAlignment}. The integrated audio-visual feature $\mathbf{z}_{cont.}$ can be computed as follows:
\begin{equation}\label{eq:gap_layer}
    \mathbf{z}_{cont.} = \frac{1}{T}\sum\nolimits_{t=1}^{T} \mathtt{concat}(\mathbf{z}_{t}^{V},\mathbf{z}_{t}^{A}),
\end{equation}

where $\mathtt{concat}$ denotes the concatenation of two vectors and $t$ denotes time step. The continuous layer feature $\mathbf{z}_{cont.}$ is obtained by temporal aggregation over all time steps $T$ by average pooling.

\paragraph{Instant Event Layer.} Another type of audio-visual event that can occur is the sparse and isolated instant ones. These actual interesting actions happen when both audio and visual signals are semantically correlated and synced for a short period as a few important moments rather than long temporal correspondence. The assigned task for this layer is performed by finding the time steps (moments) that have the highest correlation scores between audio and visual features, $\mathbf{z}^\mathbf{V}$ and $\mathbf{z}^\mathbf{A}$ respectively. \Fref{fig:layers} shows that moments with highest scores are located only in the last part of the video which can be easily understood since car appears in the scene and it is correlated with the car sound (visualized as colored frames). Remaining parts of the video are not useful for audio-visual integration as it only shows empty road like visual information. Correlation scores are computed by pairwise dot products between audio and visual embeddings~\cite{halperin2019dynamicAlignment,afouras2020AVObjects} at the same time step and audio-visual feature $\mathbf{z}_{inst.}$ is computed as follows:
\begin{equation}\label{eq:instant_layer_feature}
    \mathbf{z}_{inst.} = \frac{1}{\left|\mathcal{K}\right|}\sum_{t \in \mathcal{K}} \mathtt{concat}(\mathbf{z}_{t}^{V},\mathbf{z}_{t}^{A}), 
\end{equation}
where score list is defined as 
$\mathbf{S}_{av}[t] = \mathbf{z}_{t}^{V} {\cdot}  \mathbf{z}_{t}^{A}$
and $\mathcal{K}=\mathtt{top\textrm{-}k}(\mathbf{S}_{av})$ represents $\mathtt{top\textrm{-}k}$ sorted time steps by ${S}_{av}$. The instant layer feature is obtained by averaging the features at top-$k$ time steps. 

\paragraph{Onset Event Layer.} 
There is another type of audio-visual event that can be integrated on the event occurrences at regular points in time, \ie, rhythm~\cite{su2014avRhythm,spence2012book}. For example, sounds occur rhythmically and repetitively in dancing, musical instruments and bird calling events as they have prominent property in audio modality aligned with visual signals. Onset event layer is designed to leverage audio onsets. Audio onsets usually give information about the rhythms and beats~\cite{davis2018visualRhythm,wang2020alignnet}, musical notes and as well as beginning of the audio events~\cite{Pankajakshan2019onsetSoundEvent,Kong2018soundEvent}. As can be seen in~\Fref{fig:layers}, visual event (typebar hits the screen) occurs at the same time as onset moments which can be easily understood since this action makes rhythmic typebar sound.
Furthermore, almost equal time gap between onset moments show that this event is rhythmic. 
Audio onsets are computed by using existing audio libraries~\cite{mcfee2015librosa} for the given audio $\mathbf{A}$ as $\mathcal{O}=\mathtt{onset}(\mathbf{A})$. This return a set of time stamp indices that onsets exist in the range of $[1,T]$. We compute $\mathbf{z}_{onset}$ as follows:
\begin{equation}\label{eq:onset_layer_feature}
    \mathbf{z}_{onset} = \frac{1}{\left|\mathcal{O}\right|}\sum_{t \in \mathcal{O}} \mathtt{concat}(\mathbf{z}_{t}^{V},\mathbf{z}_{t}^{A}). 
\end{equation}

\paragraph{Visual Event Layer.} Our multisensory layers are inspired from human cognitive ability for multisensory integration as binding the multimodal signals if they are correlated and separating them otherwise. Considering some actions are soundless (``massaging legs'', ``stretching leg'', \etc) or some scenes have irrelevant sounds, integrating this irrelevant sound signals to visual features acts as noise and affects the correct prediction ability. Thus, this visual event layer is designed to recognize the events only from visual perspective. It performs the task of assigning zero-valued audio features for each visual frame feature and applying \Eref{eq:gap_layer} to output $\mathbf{z}_{visual}$.

\paragraph{Audio Event Layer.} Conversely to the visual event layer, scenes might have events that are outside of the field of view but still hearable or visual signal may be completely unrelated to the accompanying audio. Additionally, some videos might have poor visual signal. To make our network use audio only modality, this layer assigns zero-valued visual features to each audio frame feature and applies \Eref{eq:gap_layer} to compute $\mathbf{z}_{audio}$.

\vspace{-1mm}
\subsection{Training}\label{sec:training}
\vspace{-1mm}
With the proposed backbone and event-specific layers, we obtain different representations from each layer by given identical inputs, \ie audio-visual features from the backbone networks. To make each layer produces the final $C$-class prediction output ${O}_i$, letting $i$ be the index of each layer, separate fully-connected layers are used as in~\Fref{fig:pipeline}. Thus, we propose a multi-task joint learning with multiple objectives. This training methodology uses individual layers with their loss functions and supervisory label, where the same single label is given for each task. Considering this as a classification problem, cross entropy loss for each layer is computed as
\begin{equation}\label{eq:loss_layer}
\mathcal{L}_{i}({O}_{i},y)~\quad\textrm{where}~{O}_{i} = \mathtt{FC}_{i}(\mathbf{z}_i),
\end{equation}
where $i\in\{cont., inst., onset, visual, audio\}$, $\mathcal{L}$ is the cross entropy loss, $\mathtt{FC}$ is the linear classifier, ${O}$ and $y$ are the prediction output and ground-truth label respectively. The final learning objective of the network is minimizing the sum of individual losses:
\begin{equation}\label{eq:loss_function}
\mathcal{L}_{multi} = \mathcal{L}_{cont.} + \mathcal{L}_{inst.} + \mathcal{L}_{onset} + \mathcal{L}_{visual} + \mathcal{L}_{audio}
\end{equation}
where each loss has equal weight. This kind of losses (auxiliary losses) are commonly used in multi-task learning schemes and we use it for multimodal learning as in~\cite{wang2020multimodalHard}.
\vspace{-1mm}
\section{Experiments}

\subsection{Datasets}\label{sec:datasets}
We train and validate our method on five video dataset below: \textbf{VGGSound}~\cite{VGGSound} is a recently released audio-visual dataset, \textbf{Kinetics-400}~\cite{Kinetics} is a standard benchmark dataset for action recognition, \textbf{Kinetics-Sound}~\cite{arandjelovic2017L3} is a subset of Kinetics for audio-visual learning, \textbf{AVE}~\cite{tian2018ave} is audio-visual dataset formed for audio-visual event localization and \textbf{LLP}~\cite{tian2020avvp} is a multi-label dataset for audio-visual video parsing. See Supp. for details of these datasets.

\subsection{Implementation Details}\label{sec:implementation}
For all experiments, we sample audio data with 16kHz sampling rate and input audio is 10 seconds. Following previous works~\cite{afouras2020AVObjects,xiao2020avSlowFast}, we compute log-mel spectrogram with size of $1000 \times 80$. We use MC3-18~\cite{tran2018multimodalHard} as the video network and it takes 100 frames of size  $112 \times 112$ as input. Thus, $T = 100$ time steps in~\Sref{sec:backbone}. $\left|\mathcal{K}\right|$ is set to 10 for instant event layer computation. The network is trained using SGD optimizer with starting learning rate $1 \times 10^{-2}$ and reduced by a factor of 10 if validation accuracy does not increase for 3 epochs. 
See Supp.~for network architectures and audio pre-processing details.

\begin{table}[t]
{\resizebox{0.6\linewidth}{!}{
\begin{tabular}{zcccc}
\toprule
     Dataset & Uni-Audio & Uni-Vision & Naive AV  & Ours
     \\ \hline
VGGSound  &  47.0     &     40.9     &    57.1   &    \bf 59.1 \\ 
Kinetics-Sound  &  64.2      &  80.5       &   86.1         &  \bf 88.3 \\
AVE  &  79.1      &  76.1       &   86.0 &  \bf 87.8 \\
LLP  &  48.3      &  43.8       &   55.1         &  \bf 56.6 \\
Kinetics  &  21.4      &  61.0       &   66.6  &  \bf 67.0 \\
\bottomrule
\end{tabular}
}}
\caption{\textbf{Video-Level classification performance of our proposed model and baselines on five datasets.}
\vspace{-4mm}
}
\label{tab:video_level_comparison}
\end{table}

We apply the same training steps for each dataset as follows. First, we train audio backbone network from scratch with given target dataset. Then, the video backbone network is initialized by using MC3-18 trained on Kinetics-400 and fine-tuned on the target dataset. Last, we train our multi-task model with event-specific layers end-to-end by using these pre-trained backbone networks as initialization. We repeat the above step for all five datasets. Different from the others, LLP is a multi-label dataset. Since our training recipe is designed with single label, we randomly pick a label among the annotated multi-labels per sample.

\subsection{Quantitative Results}\label{sec:quantitative}
\paragraph{Video-Level Classification}\label{sec:classification}
Video classification is a task to classify a video by a single label. Since our model outputs multiple predictions from event-specific layers, we integrate them by majority voting to output a single prediction for a video-level classification task as follows:
\begin{equation}
\begin{aligned}
    \mathcal{P}_{vote} = \underset{k}{\arg\max} \sum_{i} I(p_i  = k),
\end{aligned}
\end{equation}
where $I$ is an indication function, $p_i$ is a predicted label defined as $\argmax_{j} O_{ij}$, and $j$ is an index of the vector $O_i$. In case of disagreement between the layers, that no majority consensus is obtained, the label from the most confident layer is selected.

We conduct a series of experiments to show how well our model predicts video-level labels. 
We compare the performance of our model with baselines on different datasets (\Tref{tab:video_level_comparison}). Note that our goal here is \emph{not} to compete on classification accuracy with any other expensive video recognition models, but rather to show that our model analyzes a video from different modalities and different video characteristics which leads to improvement in classification.

Accuracies on uni-modal networks in~\Tref{tab:video_level_comparison} depicts the accuracy of the backbone networks trained with single stream modalities. Naïve audio-visual model represents audio-visual networks that leverage late fusion approach (simple concat.) for final representation as used in previous works~\cite{wang2020multimodalHard,korbar2018cooperative,long2018pureAttention}. 
As results show in~\Tref{tab:video_level_comparison}, our approach offers improvement to overall performance in benchmark datasets. Our model is more effective on the datasets that are designed with audio-visual correspondence, \eg VGGSound, Kinetics-Sound, and AVE, with the improvement around ~$2\%$. Our performance improvement is less significant on Kinetics since it is a visual-dominant dataset, where many videos' sound are not correlated to visual signals.

\begin{table}[t]
{\resizebox{0.6\linewidth}{!}{
\begin{tabular}{lcccccc}
\toprule
     \multirow{2}{*}{Dataset} & \multirow{2}{*}{Ours} & \multicolumn{5}{c}{Naive AV}     \\ \cline{3-7}
     & & Top-1 & Top-2 & Top-3 & Top-4 & Top-5 \\ \hline
LLP  & \textbf{0.72} & 0.66   & 0.70    &   0.63 & 0.56 & 0.50 \\ 
\bottomrule
\end{tabular}
}}
\caption{\textbf{Multi-Label prediction measured by F1 score.}
}
\label{tab:multilabel_results}
\vspace{-0.1in}
\end{table}

\paragraph{Multi-Label Prediction}\label{sec:multilabel}
Our network contains multiple event-specific layers and each layer outputs a label prediction. The collection of each layer's prediction form a multi-label prediction set.
Let $p_i$ be a predicted label from the $i^{th}$ event-specific layer.
Then a set of label predictions $\mathcal{P}$ can be defined as follows:
\begin{equation}\label{eq:label_pred}
    \mathcal{P} = \bigcup_{i} f(p_i), \;\;\;  f(p_i) = \begin{cases}
                    p_i   &  \text{if } i = \argmax_{l} O_{lp_i}\\
                    \emptyset,              & \text{otherwise}
    \end{cases}
\end{equation}

The function $f$ accepts a predicted label $p_i$ if and only if when the value $O_{ip_i}$ is higher than the other layer values $O_{lp_i}$. This process filters out label predictions with low confidence and only highly confident predictions are gathered in a multi-label prediction set. In our implementation, we utilize five types of layers, therefore, the cardinality of the prediction set is bounded by $1 \le \vert\mathcal{P}\vert \le 5$. When our network is evaluated on the VGGSound dataset, the cardinality of $\mathcal{P}$, \ie the average number of predicted labels is $2.21$.

This analysis leads to the question of ``Are the multi-label predictions of our network are correct? or Does the VGGSound dataset contain multiple events but only annotated with a single label?''. To answer these questions, we need a multi-label evaluation. However, there is no multi-label ground truth for VGGSound dataset. Therefore, we make $12$ subjects annotate the multiple video-level predictions of our network that match with the given video for $\app$1200 videos. Note that candidate labels for annotation are limited to the labels predicted by our model. We check how many of the total predictions from our network actually do match with human answers. The user study shows that $62\%$ of all predicted labels match with human selections, which means our network on average outputs $1.4$ correct labels per sample. This study shows that VGGSound is indeed a multi-label dataset and single label prediction is insufficient to properly describe the videos.

To further evaluate the correctness of the multi-labels predicted by our network, we also conduct another experiment by using LLP as it contains multi-labels per video,~\ie the average number of labels is $1.81$. In this experiment, even though our network is trained with randomly selected single labels, it still has the ability to predict correct multi-labels for a given video. This is due to multiple layers that can output multiple contents by looking for different properties in the video. We compare our results with top-N results of Naive-AV model.
The F1 metric is used to evaluate the performance of our model and~\Tref{tab:multilabel_results} shows the results. Our model outputs multi-label predictions dynamically depending on whether the event-specific layers make consensus or not, while the top-N approach fixes the number of predictions. Although, the Top-2 performance is comparable to ours, without knowing the number of ground truth multi-labels in advance, we can not know the number of Top-N that should be selected. For example, the video shown in $2^{nd}$ row of~\Fref{fig:multilabel_vis} contains people playing more than two instruments. In this case, Top-2 only outputs two predictions while our model outputs more than two predictions. In the other case, when there is only one dominant event, our model outputs only one prediction while the Top-2 baseline still outputs two predictions.

\begin{table}[t]
{\resizebox{0.7\linewidth}{!}{
\begin{tabular}{zcccccc}
\toprule
     Dataset & Continuous & Instant & Onset & Visual & Audio & Total
     \\ \hline
VGGSound  & 62  & 39 &  68 & 41 & \textbf{99} & 309 \\
Kinetics-Sound  &  2  &  5 &  \textbf{13} & 11 & 0 & 31   \\
AVE  &  \textbf{10}  &  4 &  9 & 1 & 4 & 28   \\
LLP  &  4  &  5 &  6 & 3 & \textbf{7} & 25   \\
Kinetics  & 73    & 89    &   69 & \textbf{167} & 2 & 400 \\ 
\bottomrule
\end{tabular}
}}
\caption{\textbf{Results of modality bias on datasets.} We report the number of categories asssigned to each multisensory layer.}
\label{tab:dataset_bias}
\vspace{-0.1in}
\end{table}

\paragraph{Dataset Modality Bias}\label{sec:modality_bias}
We apply our method on understanding the modality biases of the datasets~\cite{winterbottom2020datasetBias}. Each dataset has different modality properties such as large portion of the videos may contain specific modality dominantly, \ie, Kinetics is visually heavy dataset. We pose the problem as finding the most dominant modality in a given dataset by analyzing every video. Our network outputs five different logits corresponding to the each event-specific layer. We use true class label $y$ to get confidence scores of the test video. Since this is ground truth class, the layer that is highly activated (highest confidence score for given $y$) reveals the interpretable modality information.

With this technique, we find the modality biases of the datasets as well as each category in the dataset. We apply majority voting rule within each category among all the videos and assign the modality label to the categories. See Supp. for the category-wise modality assignment.~\Tref{tab:dataset_bias} shows the number of categories that are assigned to each layer for every dataset. Our analysis shows consistent results with the prior knowledge about these benchmark datasets. Results on Kinetics clearly shows that it is visual dominant dataset. AVE dataset is curated for audio-visual learning and our method validates this as one of the AV layers is dominant. LLP~\cite{tian2020avvp} reports that majority of the annotated events are audio events. Our analysis also confirms that LLP has tendency to the audio modality.  

Additionally, we perform an experiment to see how many categories of Kinetics-Sound (created by selecting categories manually from Kinetics in~\cite{arandjelovic2017L3}) matches with the categories that our audio-visual event-specific layers filter in Kinetics. This reveals that $66\%$ of the Kinetics-Sound categories are matched. Thus, our modality bias selection gives consistent results with human selections. Please see Supp. for details.

\begin{table}[t]
{\resizebox{0.8\linewidth}{!}{
\begin{tabular}{zcccc}
\toprule
     Dataset & Continuous & Instant & Onset  & (Ins.$\lor$Ons.) $\land$ ($\lnot$Cont.)
     \\ \hline
VGGSound  & 354  & 407 &  230 & 778 \\
Kinetics-Sound  &  18  &  18 &  9 & 34   \\
AVE  &  7  &  4 &  3 & 12   \\ 
LLP  &  25  &  47 &  22 & 74   \\
Kinetics  & 366    & 567    &   258 & 985 \\ 
\bottomrule
\end{tabular}
}}

\caption{\textbf{Layer-wise statistics of correctly predicted videos.
}
}
\label{tab:layer_diff}
\vspace{-0.1in}
\end{table}

\paragraph{Layer Differences}\label{sec:layer_diff}
At first glance, the differences between each audio-visual event layer may not look significant. 
However, they indeed look for different audio-visual characteristics in the videos. 
We count the number of unique videos that one of these layers classifies correctly while the rest two layers fail, and also the samples that continuous layer fails but instant or onset layer predict correctly in \Tref{tab:layer_diff}. The number of uniquely correct samples for each layer is comparable to each other and the last column shows that the instant and onset layers capture significant amount of correct samples compared to the continuous layer. This shows that the proposed multisensory layers are complementary. 
As a consequence, the model enables a more in-depth interpretation of videos.

Continuous event layer pools information from every time step features of both modality without consideration of their correspondence. Therefore, number of the informative features may be less than uninformative ones for some videos, as in ``car'' example of~\Fref{fig:layers}. Thus, using informative subset of these every time step features, \ie, instant or onset, may improve the accuracy for these videos since it will be less contaminated. Furthermore, the difference between instant and onset event layer can be also seen in the ``car'' example of~\Fref{fig:layers}. Onset event layer uses the onset moments, \ie pink dots, that are grouped in the part before the real action starts whereas instant event layer captures the instant, highly audio-visual correlated moments of the video, \ie blue dotted box.

\subsection{Qualitative Results}\label{sec:qualitative}
\noindent\textbf{Audio-Visual Attention.} As aforementioned in~\Sref{sec:task_layers}, instant event layer is leveraged to use highly correlated audio-visual moments. Thus, to gain a better understanding of these moments, we visualize the localization response $\balpha$ where $\balpha = {\mathcal{V}_t}{\cdot}  \mathbf{z}_{t}^{A}$, ${\mathcal{V}_t} {\in} \mathbb{R}^{H' {\times} W' {\times} D}$ is the visual activation from the last convolution layer of video backbone network and $\mathbf{z}_{t}^{A}$ is the audio embedding at moment $t$~\cite{senocak2019learningTPAMI,afouras2020AVObjects}.

\begin{figure}[t]
\centering
		\includegraphics[width=0.91\linewidth]{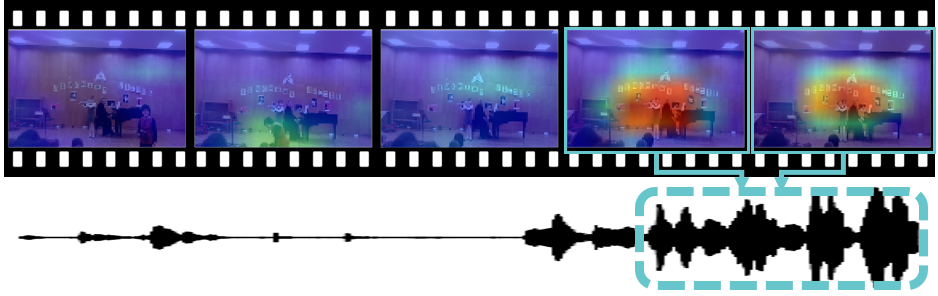}
\caption{\textbf{Sound Localization.} Our backbone networks correctly localize the sound spatially and time-wise as a natural outcome of the model without any explicit training for this task.\vspace{-3mm}}
\label{fig:localization_vis}
\end{figure}

We qualitatively show in~\Fref{fig:localization_vis} that features from our backbone networks can locate the sound source without any separate training mechanism for this task. The localization response only activates when the girl plays flute, otherwise inactivated. This confirms that our model not only attends where the source appear in the video spatially but also attends when the event sound occurs time-wise.

We perform an additional experiment to show the potential of our model quantitatively. 
We use a recently released VGG-SS dataset\cite{chen2021localizing} which is curated from VGGSound that contains around 5K samples. Following the same setup as~\cite{chen2021localizing}, we compute the 3 sec.~time average of localization responses around mid-frame. As shown in~\Tref{tab:localization_performance}, even though our method is not trained with the sound localization objective, it still achieves competitive results to explicitly trained state-of-the-art sound localization method~\cite{chen2021localizing}.

\begin{table}[t]
{\resizebox{0.5\linewidth}{!}{
\tiny
\begin{tabular}{z@{\hspace{0.4cm}}c@{\hspace{0.4cm}}c}
\toprule
     Method & IoU & AUC \\ \hline
Attention.~\cite{senocak2018learning}    &  0.185 & 0.302  \\
AVEL~\cite{tian2018ave}     		       & 0.291 & 0.348 \\
AVobject~\cite{afouras2020AVObjects}     		       & 0.297 & 0.357 \\
LVS~\cite{chen2021localizing}$\dagger$    		       & 0.303 &  \textbf{0.364} \\
\textbf{Ours}         					     & \textbf{0.309} & 0.354 \\
\bottomrule
\end{tabular}
}}
\caption{\textbf{Quantitative sound localization results on the VGG-SS~\cite{chen2021localizing} test set}. All models are trained on VGG-Sound 144k and tested on VGG-SS.  $\dagger$ is the result of the model released on the official project page and the authors report ~3$\%$ drop in IoU performance comparing to their paper.}
\label{tab:localization_performance}
\end{table}

\begin{figure}[!t]
\centering
\small
{
\resizebox{.7\linewidth}{!}{%
\begin{tabular}{c}

\includegraphics[width = .61\linewidth]{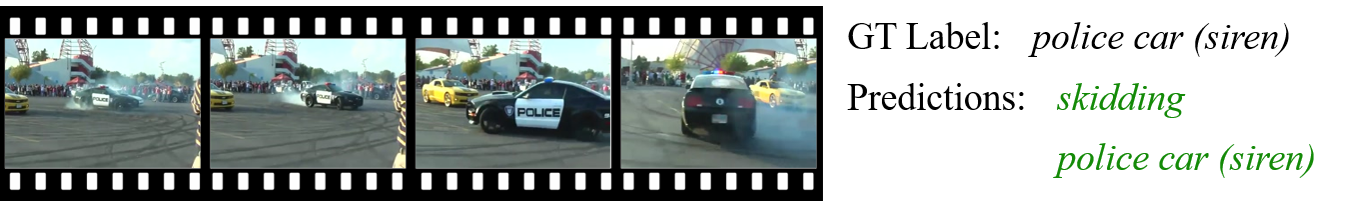} \\
\includegraphics[width = .61\linewidth]{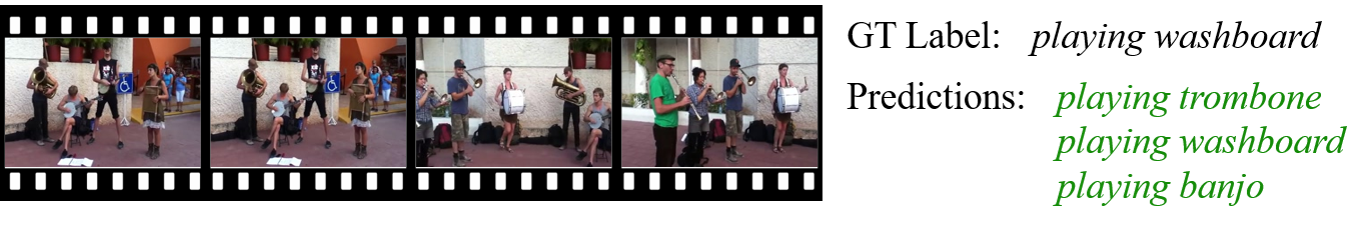} \\
\includegraphics[width = .61\linewidth]{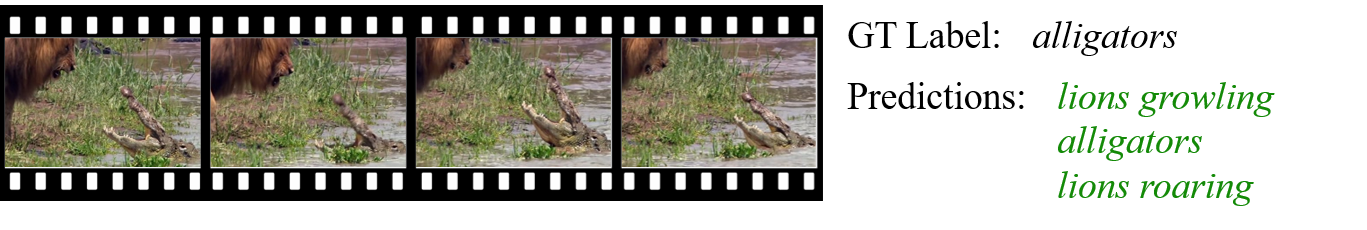} \\
\includegraphics[width = .61\linewidth]{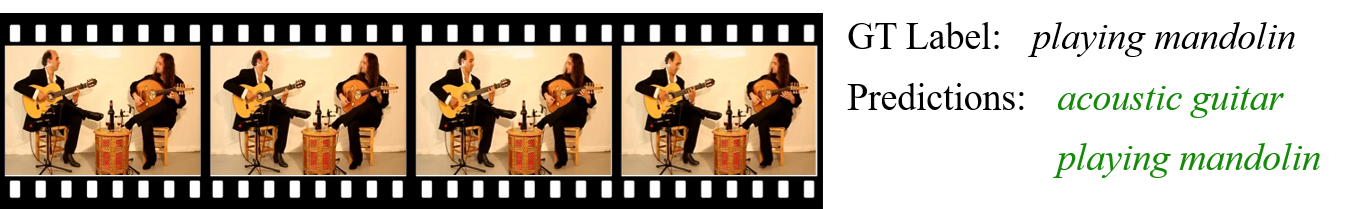} \\
\end{tabular}
}
}
\caption{\textbf{Single- to Multi-Label Prediction.} The original annotations are single labels , whereas the videos contain multiple events, actions, or categories. 
Our method predicts true multi-labels that leads to more accurate video labels.\vspace{-2mm}}
\label{fig:multilabel_vis}
\vspace{-4mm}
\end{figure}

\paragraph{Multi-Labeling Videos.} We qualitatively show the multi-labeling ability of our network even though the network is trained on single labels.~\Fref{fig:multilabel_vis} shows predicted multi-labels are consistent with the human perception.  

\begin{figure*}[!t]
	\begin{center}
		\includegraphics[width=1\linewidth]{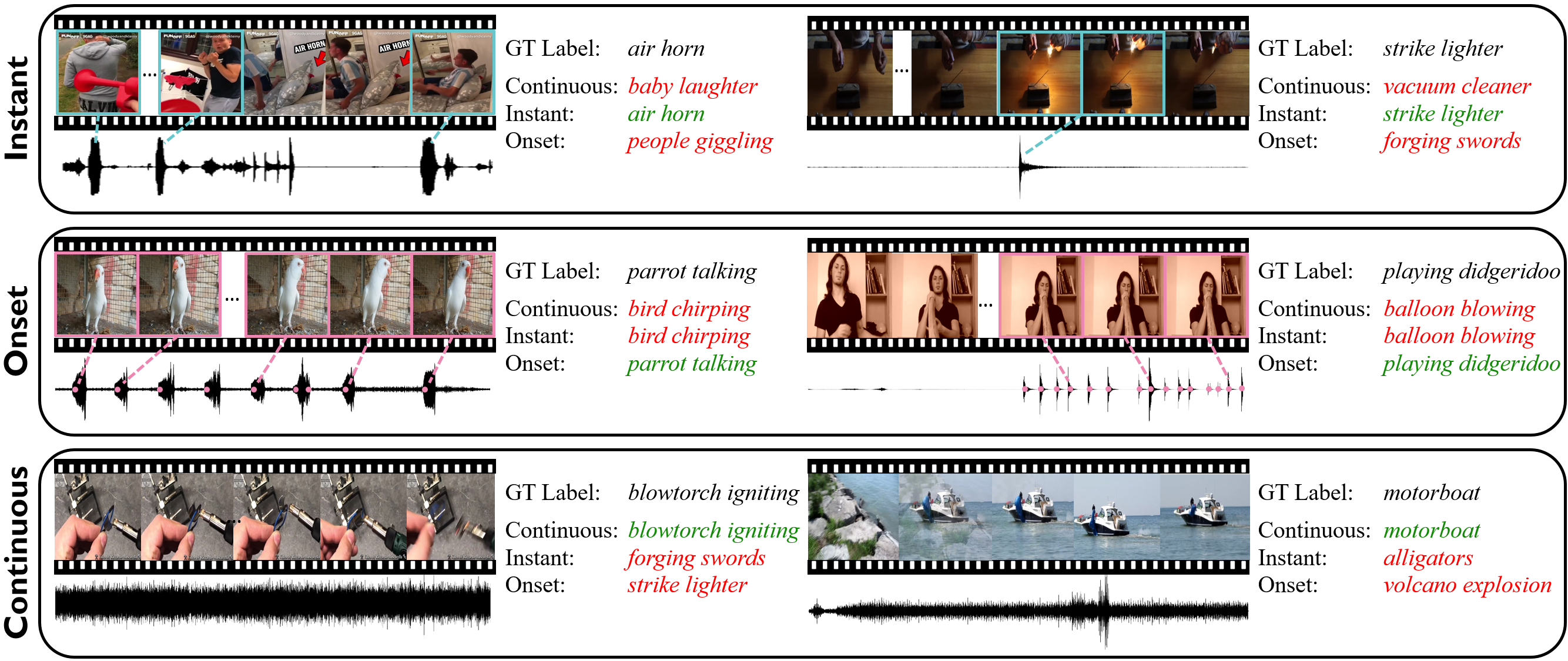}
	\end{center}
	\caption{\textbf{Visualization of the representative characteristics from event-specific layers.} 
	Each layer captures distinctive audio-visual characteristics.
	Note that our multisensory model not only correctly detects the event types, but also makes accurate category prediction within the layers.}
	\label{fig:layer_vis}
\vspace{-4mm}
\end{figure*}

\paragraph{Layer Visualizations.} In order to have a better insight on what each event-specific layer learned, we visualize some of the videos that are maximally activated by a certain layer.~\Fref{fig:layer_vis} displays some videos for each audio-visual layer. As it can be seen, instant event layer has short period high intensity-like patterns; onset event layer captures rhythmic-like patterns; and continuous event layer shows temporally constant-like events. Note that these patterns show similarities to abstract event visualizations in~\Fref{fig:second_teaser}. We also visualize that representations from these assigned layers may be more proper for given videos as only that certain layer makes correct prediction. 


\section{Concluding Remarks}
\noindent
We present a multisensory model with event-specific layers that incorporates different audio-visual relationships and demonstrate the efficacy of our model on five different video datasets with a diverse set of videos. Unlike previous audio-visual models, our event-specific layers output multiple predictions with modality confidences. This leads to new future research directions for audio-visual understanding. We propose interesting potential applications that can be built based on our model and discuss the limitations of our work.

\paragraph{Limitations.} Our system can predict a multiple labels with confidence. Yet this raw predictions is not normalized. In other words, a prediction with highest confidence is not necessarily more dominant or accurate than the other predictions. A proper calibration of the confidences will enable our model to select a most prominent event in a video. 
Another limitation is false positives in multi-label predictions. Although majority of our model's multi-label prediction aligns with human perception as in~\Tref{tab:multilabel_results} and ~\Fref{fig:multilabel_vis}, there is still gap in the quantitative results. One possible future direction is to design a prediction fusion mechanism to keep semantically meaningful predictions while removing false predictions.

\vspace{2mm}
 
\paragraph{Modality-level Video Understanding.}
Even for a single video clip, at different timestamps, different modality might play the key role in video understanding.
For example, a part of the video may not be informative due to  motion blur or occlusion and the audio modality may be more indicative of what is happening.
Similarly, video can be more indicative modality when the microphone is accidentally blocked. Our method can tell which modality to rely to understand ongoing events in a video using a confidence from each layer. 
This kind of modality-level video understanding, as oppose to category-level, is crucial and complementary to existing methods.
One can use our network to decide which modality is playing the key role and apply the right specialized inference network to get the actual predictions.

\paragraph{Missing Label Detection.}
Given single label datasets as input, we demonstrate that our system can discover accurate multiple labels automatically (See \Fref{fig:multilabel_vis}). 
In the manual annotation process, it is common to only label the most salient objects or events. 
The future of datasets should have more detailed labels, beyond the sole event in the scene. 
For example, in \Fref{fig:multilabel_vis} ``alligators'' video, only the label ``alligators'' is annotated.
However, clearly there are other audio-visual contents such as lions growling, lions roaring, and water splashing sound by the alligator present in the video that are not properly annotated.
Our system can automatically discover these potential labels in the same video and therefore build up more comprehensive dataset.

\paragraph{Dataset Retargeting / Cleanup.}
Our work could be used to pre-screen any video dataset to curate specific modalities that are of interests to training.
Datasets, especially those constructed from user-uploaded YouTube videos, are highly biased towards certain modalities (see Table~\ref{tab:dataset_bias}).
If one tries to use Kinetics dataset for an audio-oriented training, the outcome would be bad since only 2 out of 400 categories are audio-focused, according to our analysis. 
Our proposed method can identify the modality distribution of dataset and only select those videos that match the requirements of the target application. For example, sound localization requires strong correspondences between audio and visual domain, video classification may be sufficient with strong visual information like Kinetics. More specifically, we can select only the audiovisual videos from Kinetics and construct a Kinetics-AV which would have a high efficacy in audiovisual learning tasks. Combining with the label prediction from those modalities, we foresee a significantly easier process to create application-specific datasets in the future.




%
%
\bibliographystyle{splncs04}
\bibliography{egbib}

\newpage
\noindent {\LARGE \textbf{Appendix}}
\section*{Supplementary Material}
In this supplementary material, we present additional details pertaining to the experiments that are not included in the main text due to space constraints. All figures and references in this supplementary material are self-contained.

The contents included in this supplementary material are as follows: 1) Details of the backbone network architectures, audio pre-processing, and datasets, 2) Layer analysis with VGGSound categories, 3) Additional sound localization results, 4) Potential applications.

\setcounter{section}{0}
\section{Architecture Details of Backbone Networks}\label{sec:architecture}
In~\Tref{tab:arch}, we provide the architecture of the backbone networks. We use two-stream network architecture, video network and audio network, as in existing audio-visual learning works. The video network is a spatio-temporal ResNet mixed convolution network, similar to MC$x$~\cite{tran2018multimodalHard}, borrowed from official PyTorch implementation\footnote{https://pytorch.org/vision/0.8/models.html\#torchvision.models.video.mc3\_18} (mc3\_18). Audio network is a custom network built with 2D convolution layers.  Batch Normalization and ReLU activation function are used after every convolution layer.

\begin{table}[h!]
\centering
\caption{\textbf{Architecture of the backbone networks.} $K$, $S$, $P$, $res$, $max pool$ and $avg pool$ denote kernel size, stride, padding, residual, max-pooling and average-pooling layers, respectively.}
\scriptsize
\begin{tabular}[t]{  l r c c c r }
 \toprule
 Layer & \# filters & K & S & P   & Output  \\  
 \midrule
 input  & 1 &   - &  -  & -           & $10 \times 100 \times 80 $  \\  
 conv1  & 64 & (1,3,3)   & (1,2,1) & (0,1,1)    & $10 \times 50 \times 80 $  \\  
 conv2 & 64 & (1,3,3)   & (1,1,2)& (0,1,1)     & $10 \times 50 \times 40 $  \\ 
 maxpool2   & -   & (1,1,3)   & (1,1,2)& (0,0,0)     & $10 \times 50 \times 19 $  \\ 
 conv3 & 192 & (1,3,3)   & (1,1,1) & (0,1,1)    & $10 \times 50 \times 19 $  \\
 maxpool3   & -   & (1,3,3)   & (1,2,2)& (0,0,0)     & $10 \times 24 \times 9 $  \\
 conv4 & 256 & (1,3,3)   & (1,1,1)& (0,1,1)     & $10 \times 24 \times 9 $  \\
 conv5 & 256 & (1,3,3)   & (1,1,1)& (0,1,1)     & $10 \times 24 \times 9 $  \\
 conv6 & 256 & (1,3,3)   & (1,1,1)& (0,1,1)     & $10 \times 24 \times 9 $  \\
 maxpool6   & -   & (1,3,2)   & (1,2,2)& (0,0,0)     & $10 \times 11 \times 4 $  \\
 conv7 & 512 & (1,4,4)   & (1,1,1)& (0,1,0)     & $10 \times 10 \times 1 $  \\
 fc8   & 512  & (1,1,1)   & (1,1,1)& (0,0,0)     & $100 \times 1$  \\
 fc9   & 512 & (1,1,1)   & (1,1,1)& (0,0,0)     & $100 \times 1$  \\
 \bottomrule
\end{tabular}
\subcaption*{{\bf (a)} Audio Network}

\begin{tabular}[t]{  l r r r r r r }
 \toprule
 Layer & \# filters & K & S & P  & Output  \\  
 \midrule
 input  & 3 &   - &  -  & -                 & $100 \times 112 \times 112$  \\  
 conv1    & 64 & (3,7,7) & (1,2,2) & (1,3,3)   & $100 \times 56 \times 56$  \\  
 conv2 & 64 & (3,3,3)   & (1,1,1) & (1,1,1)     & $100 \times 56 \times 56$  \\ 
 conv3 & 64 & (3,3,3)   & (1,1,1) & (1,1,1)     & $100 \times 56 \times 56$  \\ 
 conv4 & 64 & (3,3,3)   & (1,1,1) & (1,1,1)     & $100 \times 56 \times 56$  \\
 conv5 & 64 & (3,3,3)   & (1,1,1) & (1,1,1)     & $100 \times 56 \times 56$  \\
 conv6 & 128 & (1,3,3)   & (1,2,2) & (0,1,1)     & $100 \times 28 \times 28$  \\
 conv7  & 128  & (1,3,3)   & (1,1,1) & (0,1,1)     & $100 \times 28 \times 28$  \\
 res-conv8  & 128 & (1,1,1)   & (1,2,2) & (0,0,0)     & $100 \times 28 \times 28$  \\
 conv9  & 128  & (1,3,3)   & (1,1,1) & (0,1,1)     & $100 \times 28 \times 28$  \\
 conv10  & 128  & (1,3,3)   & (1,1,1) & (0,1,1)     & $100 \times 28 \times 28$  \\
 conv11  & 256  & (1,3,3)   & (1,2,2) & (0,1,1)     & $100 \times 14 \times 14$  \\
 conv12  & 256  & (1,3,3)   & (1,1,1) & (0,1,1)     & $100 \times 14 \times 14$  \\
 res-conv13  & 256  & (1,1,1)   & (1,2,2) & (0,0,0)     & $100 \times 14 \times 14$  \\
 conv14  & 256  & (1,3,3)   & (1,1,1) & (0,1,1)     & $100 \times 14 \times 14$  \\
 conv15  & 256  & (1,3,3)   & (1,1,1) & (0,1,1)     & $100 \times 14 \times 14$  \\
 conv16  & 512  & (1,3,3)   & (1,2,2) & (0,1,1)     & $100 \times 7 \times 7$  \\
 conv17  & 512  & (1,3,3)   & (1,1,1) & (0,1,1)     & $100 \times 7 \times 7$  \\
 res-conv18  & 512  & (1,1,1)   & (1,2,2) & (0,0,0)     & $100 \times 7 \times 7$  \\
 conv19  & 512  & (1,3,3)   & (1,1,1) & (0,1,1)     & $100 \times 7 \times 7$  \\
 conv20  & 512  & (1,3,3)   & (1,1,1) & (0,1,1)     & $100 \times 7 \times 7$  \\
 avgpool   & -   & (1,7,7)   & - & (0,0,0)     & $100 \times 1 \times 1$  \\
 \bottomrule
\end{tabular}
\subcaption*{{\bf (b)} Video Network}
\label{tab:arch}
\end{table}

\begin{table}
\tiny
{\resizebox{0.6\linewidth}{!}{
\begin{tabular}{lccccc}
\toprule
Dataset & Train & Test & Val.  & Total
     \\ \hline
VGGSound  & 170384  & 0 &  13675 & 184059 \\ 
Kinetics  & 208552    & 33595    &   17019 & 259166 \\ 
Kinetics-Sound  &  19931  &  2677 &  1351 & 23959   \\
AVE  &  3697  &  402 &  0 & 4099   \\
LLP  &  9620  &  1162 &  624 & 11406   \\
\bottomrule
\end{tabular}
}}

\caption{\textbf{Dataset statistics in our experiments.}
\vspace{-1mm}
}
\label{tab:dataset_diff}
\end{table}

\begin{figure*}[!t]
	\begin{center}
		\includegraphics[width=1\linewidth]{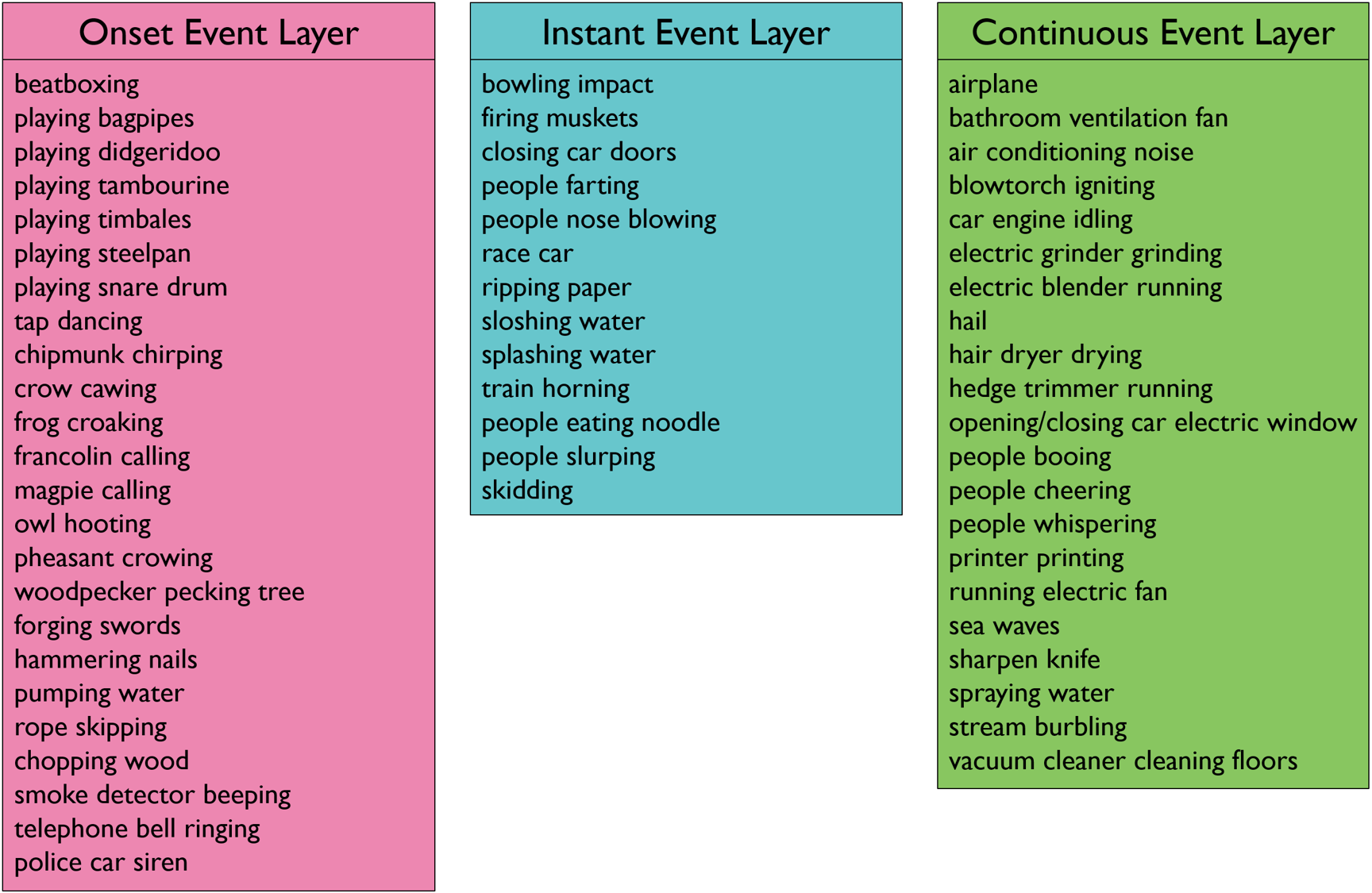}
	\end{center}
	\caption{\textbf{VGGSound categories that are assigned to each audio-visual event layer.}}
	\label{fig:layer_category}
\end{figure*}

\section{Audio Pre-processing Details}\label{sec:audio_preprocess}
We sample audio data with 16kHz sampling rate and input audio is 10 seconds. STFT is computed using $n\_fft=512$, $hop\_length=160$, $win\_length=320$, $window=hann\_window(320)$, $center=True$ and $pad\_mode=reflect$ and $1000 \times 80$ log-mel spectrogram is produced with 80 mel filterbanks by using PyTorch. Audio onsets are computed using librosa~\cite{mcfee2015librosa} onset detection function with a pre-computed onset envelopes. 
\section{Datasets}\label{sec:datasets_details}
We train and validate our method on five video datasets using standard evaluation metrics. \textbf{VGGSound}~\cite{VGGSound} is a recently released audio-visual dataset which contains around $\app$200K videos obtained from YouTube and labelled with 309 categories. \textbf{Kinetics-400}~\cite{Kinetics} is a large-scale standard benchmark dataset for action recognition with $\app$240K training and 20K validation videos containing 400 human action classes. \textbf{Kinetics-Sound}~\cite{arandjelovic2017L3} is created by choosing 34 classes from Kinetics dataset that are assumed to have audio and visual characteristics and it has total $\app$22k videos. \textbf{AVE}~\cite{tian2018ave} is another audio-visual dataset formed for audio-visual event localization and it contains around 4K videos covering 28 event categories. \textbf{LLP}~\cite{tian2020avvp} is a multi-label dataset consisting of $\app$12k videos labeled by 25 categories and formed for audio-visual video parsing.

However, some of the videos are removed or not accessible from the web because of privacy or regional settings. Hence, our datasets may be slightly smaller than official numbers for some datasets (See~\Tref{tab:dataset_diff}). Additionally, the original 34 classes in Kinetics-Sound are based on the earlier version of the Kinetics. Some classes are removed currently. Therefore, we use available 31 classes.

\section{Category-wise Layer Analysis}\label{sec:layer_categories}
In~\Fref{fig:layer_category}, we list the VGGSound categories that are assigned to each Audio-Visual event-specific layer of our network. 
To obtain these results, we apply majority voting rule among all the videos within each category and assign the layer label to the categories as we explain in ``Dataset Modality Bias'' subsection of the main paper. We show some of these categories (due to the limited space) for each audio-visual event layers in~\Fref{fig:layer_category}. 

The onset event layer predicts categories that are related to music, animal and repetitive actions or sounds as shown in~\Fref{fig:layer_category}. As aforementioned in the main paper, musical instruments such as ``playing tambourine'', ``playing steelpan'', ``beatboxing'' or animal vocalization sounds -- ``frog croaking'', ``francolin calling'', ``chipmunk chirping'' -- and some actions such as ``hammering nails'', ``forging swords'', ``smoke detector beeping'' all tend to have rhythmic and repetitive characteristics. This aligns with our design motivation that onset event layer learns rhythmic, repetitive and periodic events as listed categories contain these characteristics.

The instant event layer focuses on the categories that contain impact events like ``bowling impact'', ``closing car doors'' or sudden events like ``people nose blowing'', ``train horning'' or explosion-kind of events such as ``firing muskets'', ``splashing water'' and ``people farting''. This also matches with our intuition that instant event layer predicts sudden, sparse highly audio-visual correlated instant events.

Finally, the categories that are highlighted by the continuous event layer have temporally constant-like characteristics such as ``bathroom ventilation fan'', ``blowtorch igniting'', ``hair dryer drying'' or slowly evolving sounds like ``airplane'' or ``sea waves''. The results also show that our intuition on the continuous event layer matches with these categories.


\begin{figure*}[h]
\centering
{
\resizebox{0.8\linewidth}{!}{%
\begin{tabular}{c}
\includegraphics[width = 1\linewidth]{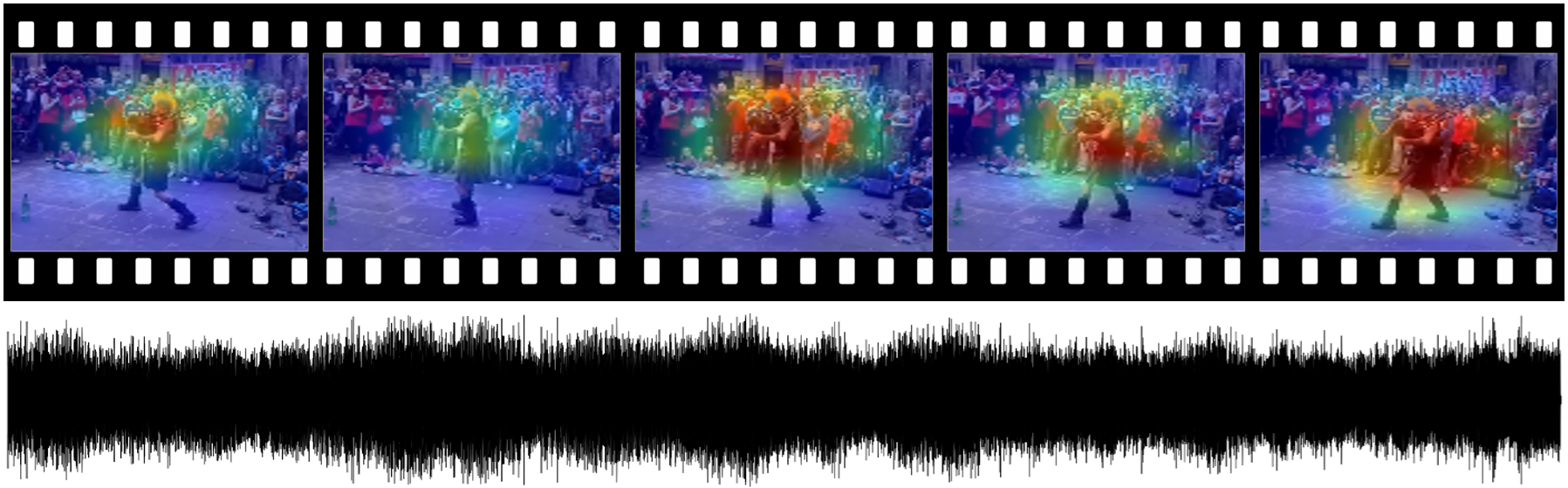} \\
\includegraphics[width = 1\linewidth]{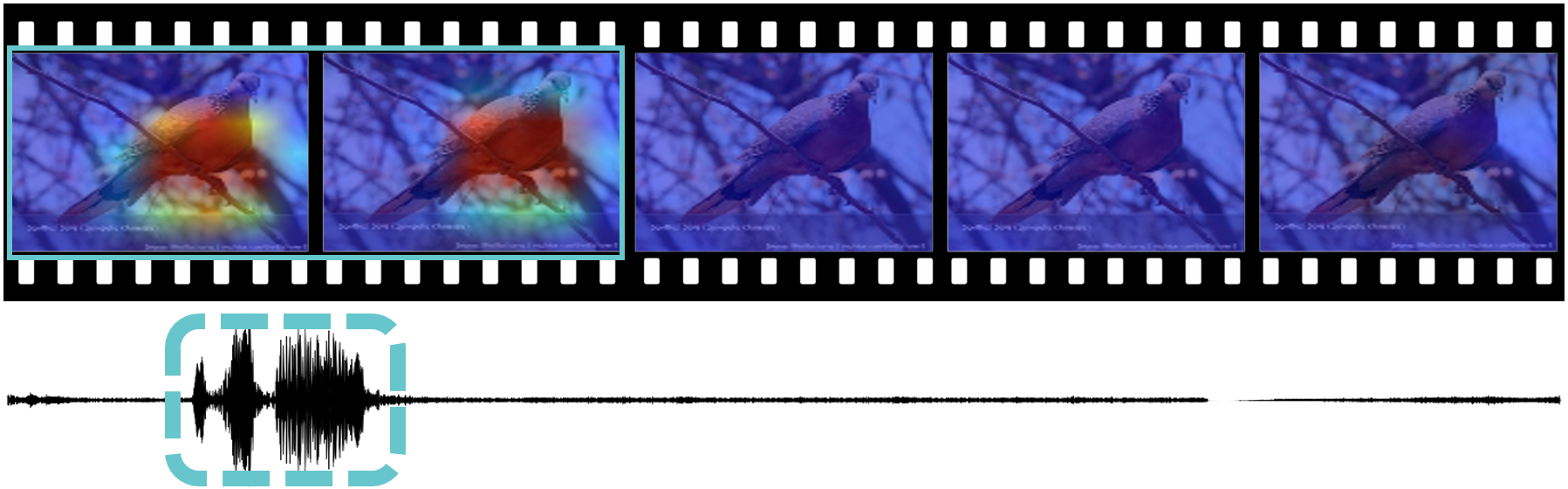} \\
\includegraphics[width = 1\linewidth]{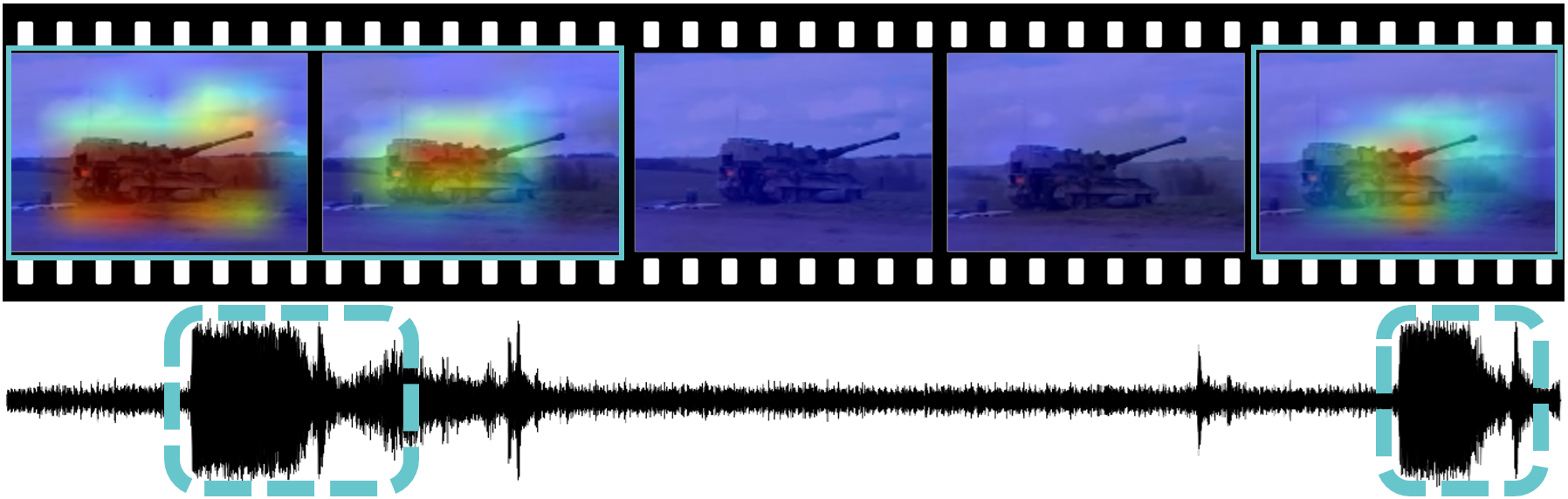} \\
\includegraphics[width = 1\linewidth]{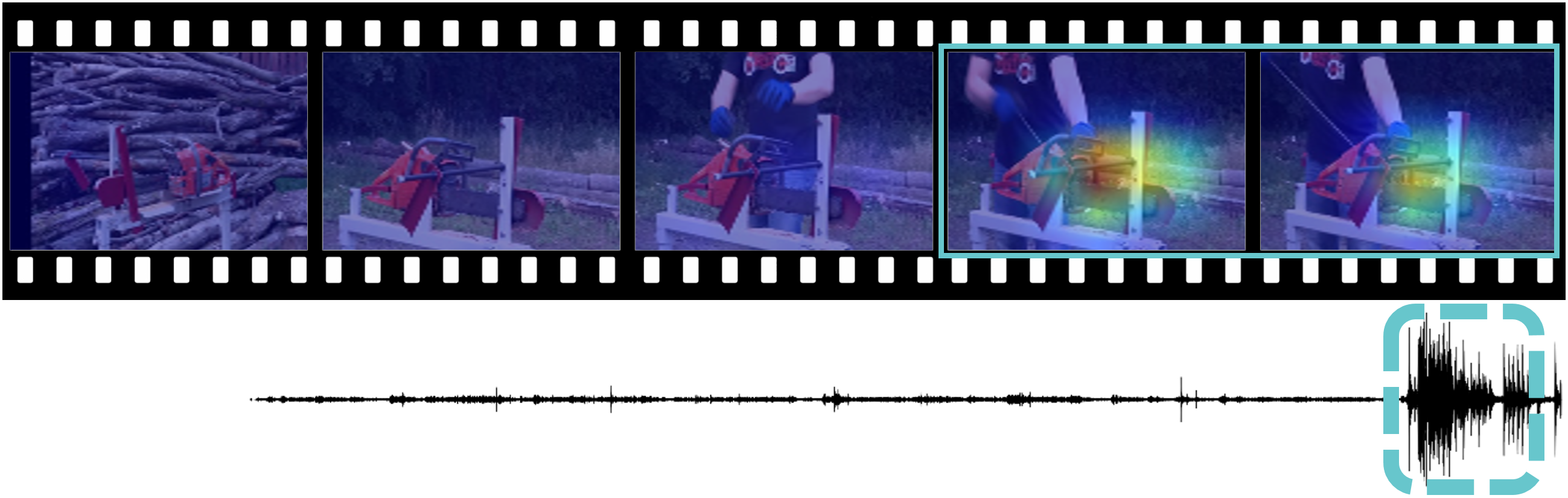} \\
\end{tabular}
}
}
\caption{\textbf{Additional Sound Localization Results.}\vspace{-2mm}}
\label{fig:sound_localization}
\vspace{-4mm}
\end{figure*}

\section{Qualitative Results of Sound Localization}\label{sec:sound_localization}
~\Fref{fig:sound_localization} shows additional qualitative results of the sound localization attempt, spatially and time-wise, by using the features from our backbone networks throughout videos. 

\begin{figure*}[h]
\centering
{
\resizebox{0.8\linewidth}{!}{%
\begin{tabular}{c}
\includegraphics[width = 1\linewidth]{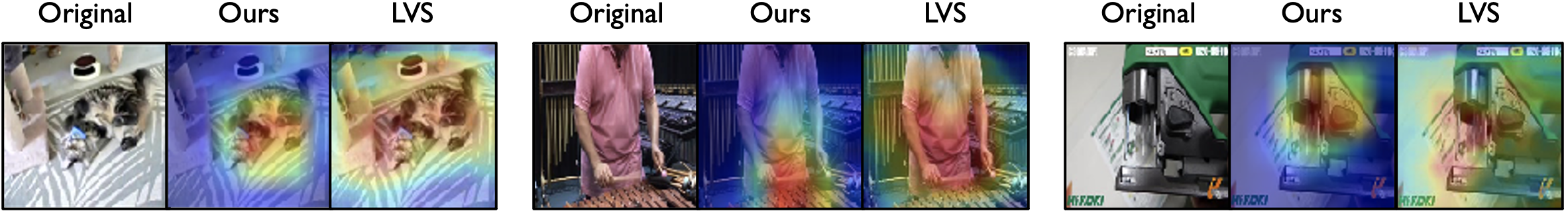} \\
\includegraphics[width = 1\linewidth]{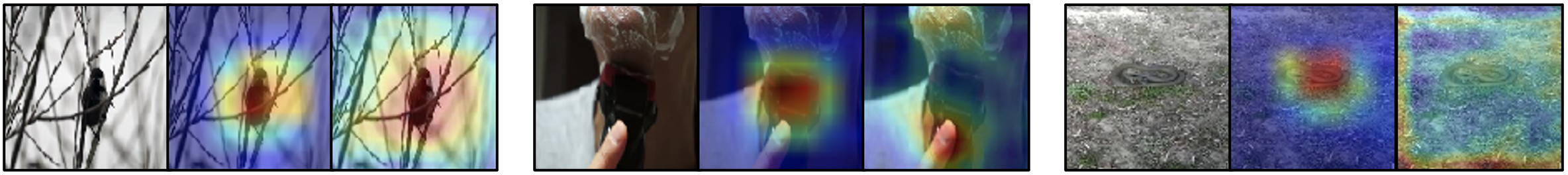} \\
\includegraphics[width = 1\linewidth]{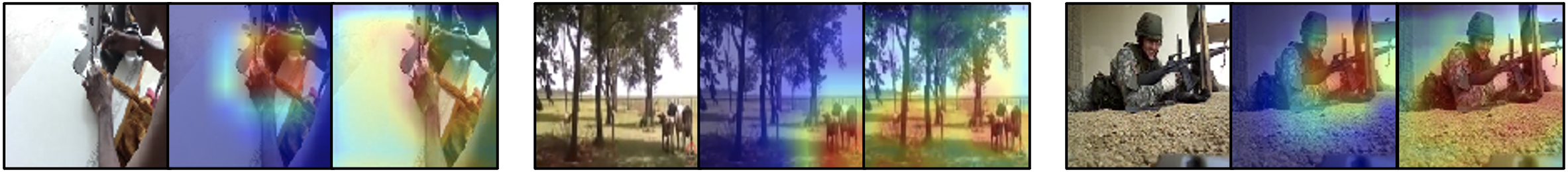} \\
\includegraphics[width = 1\linewidth]{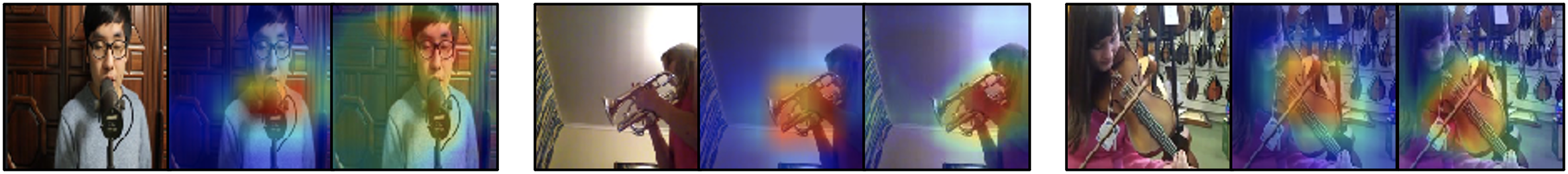} \\
\includegraphics[width = 1\linewidth]{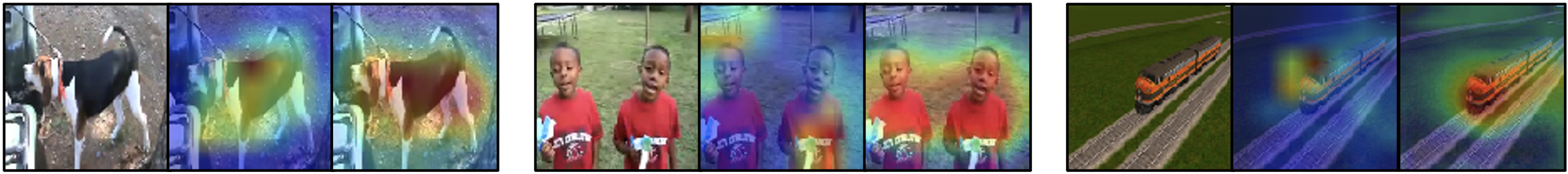} \\
\end{tabular}
}
}
\caption{\textbf{Sound Localization Results on VGG-SS and comparison with LVS~\cite{chen2021localizing}.}}
\label{fig:qualitative_vggss}
\vspace{-4mm}
\end{figure*}

We also visualize the attention maps of VGG-SS test samples in~\Fref{fig:qualitative_vggss} and compare them with the state- of-the-art~\cite{chen2021localizing} method on this dataset. This figure shows how two different approaches response to the same samples. 

\section{Potential Applications}\label{sec:applications}
In ``Concluding Remarks'' section of the main paper, we have discussed potential applications that can be built based on our model. In this supplementary material, we present some examples for these applications.

\subsection{Dataset Retargeting / Cleanup}
\begin{figure*}[h]
	\begin{center}
		\includegraphics[width=1\linewidth]{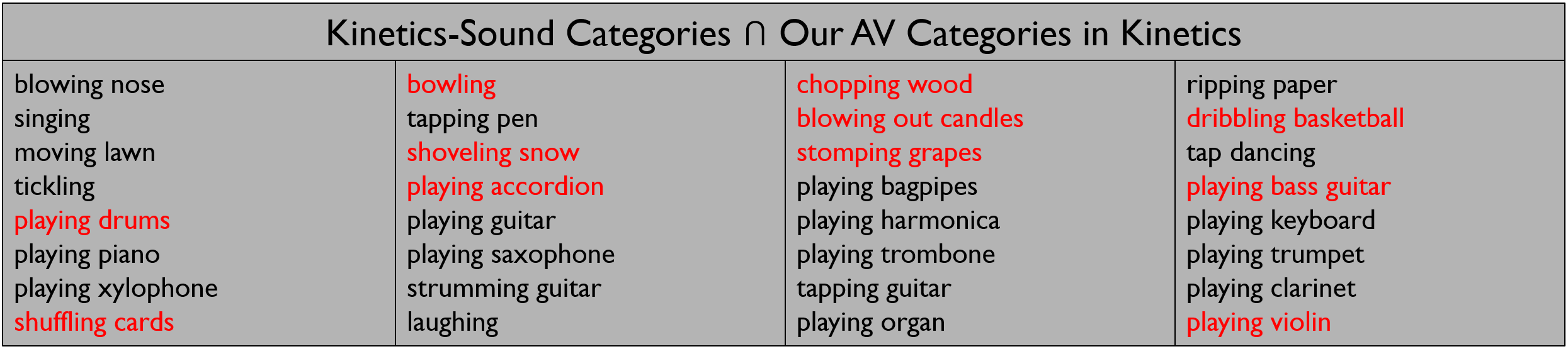}
	\end{center}
	\caption{\textbf{Kinetics-Sound categories that intersect with our AV categories in Kinetics.}}
	\label{fig:kinetics_sound}
\end{figure*}

As an example to dataset retargeting application, we use our proposed method to identify the modality distribution of Kinetics dataset and only select the categories that fall under the audio-visual event layers. Kinetics-Sound dataset is constructed to have a high efficacy in audio-visual learning tasks. Based on the subset we construct, we perform an experiment to see how many categories of Kinetics-Sound matches with the categories that our audio-visual event layers filter. This experiment reveals that $66\%$ of the Kinetics-Sound categories are matched. Kinetics-Sound categories that intersect with the categories that our audio-visual event layers filtered from Kinetics are listed in~\Fref{fig:kinetics_sound}.

\subsection{Modality-level Video Understanding}
\begin{figure*}[h]
	\begin{center}
		\includegraphics[width=1\linewidth]{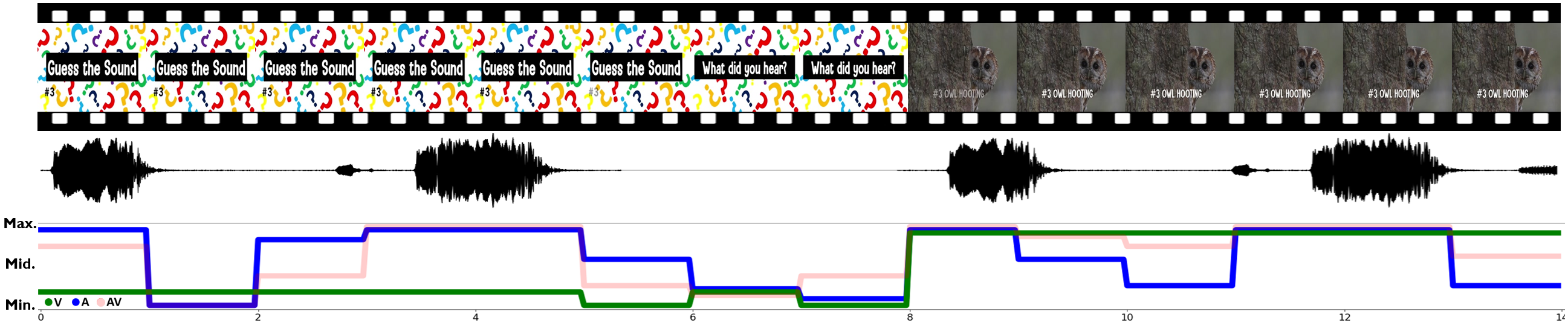}\vspace{-3mm}
	\end{center}
	\caption{{\bf Modality-level Video Analysis Results.} Plotted lines at the bottom of the figure depicts the modality confidence level throughout the video according to the informative signals.}
	\label{fig:modality_analysis}
	\vspace{-2mm}
\end{figure*}

We present an example of modality-level video understanding in~\Fref{fig:modality_analysis}. 
As shown in the figure, the modality confidence of each layer is highly activated only when there is meaningful signal.
The audio modality confidence level aligns with the audio signal presence.
Similarly, vision modality confidence is activated when the meaningful frames, \ie, frames with owl, appear.
The audio-visual modality, which is per second continuous layer in this example, is activated when either modality is confident.

\subsection{Missing Label Detection}
\begin{figure*}[h]
\centering
{
\resizebox{1\linewidth}{!}{%
\begin{tabular}{c}
\includegraphics[width = 1\linewidth]{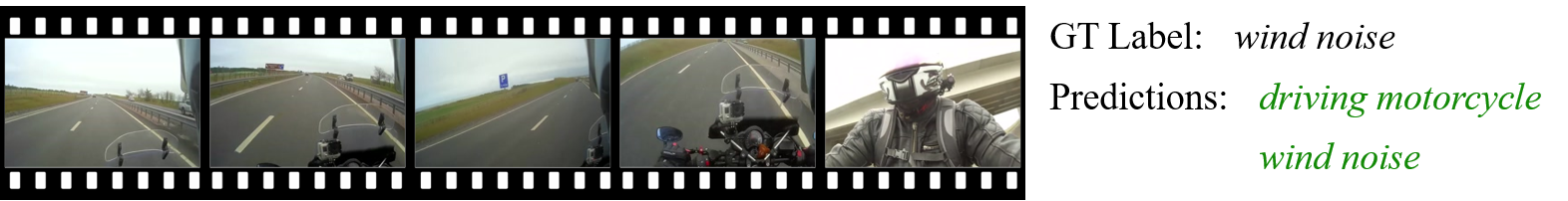} \\
\includegraphics[width = 1\linewidth]{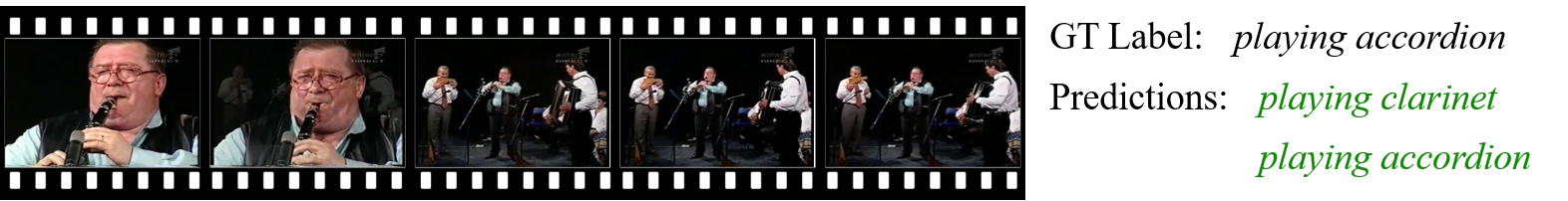} \\
\includegraphics[width = 1\linewidth]{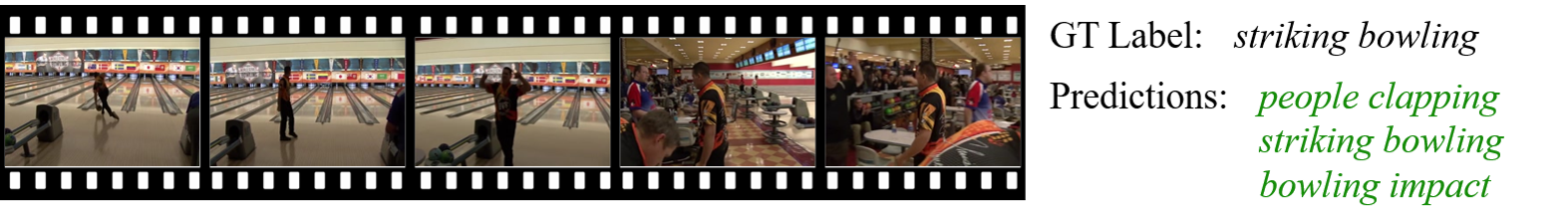} \\
\includegraphics[width = 1\linewidth]{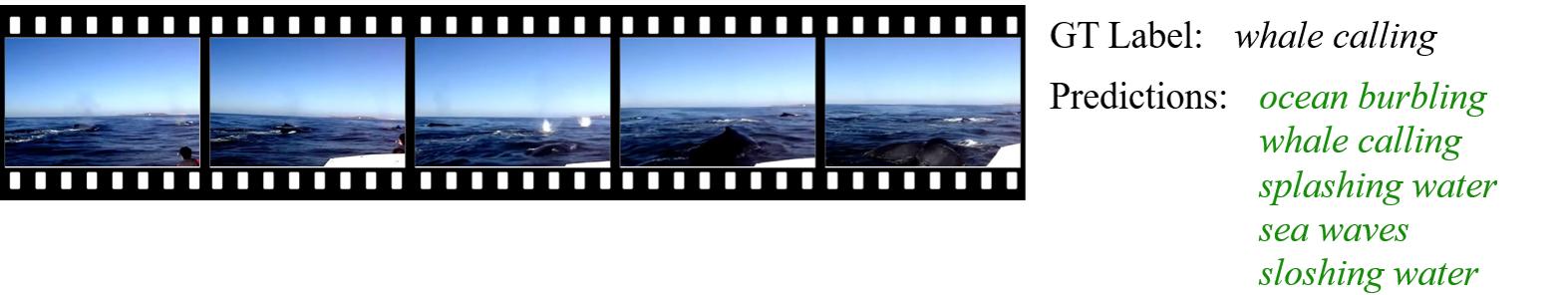} \\
\end{tabular}
}
}
\caption{\textbf{Missing Label Prediction.} Our method predicts additional labels that are contained but not labeled in the given video examples.\vspace{-2mm}}
\label{fig:missinglabel_vis}
\vspace{-4mm}
\end{figure*}

Often, a video contains multiple objects and events with complex interactions. It is difficult to fully represent contents of a video with a single label. Moreover, humans make mistakes during annotations by missing annotation. Our model outputs multiple labels using event-specific layers which capture different characteristics of a video. We provide multi-label prediction examples using videos from VGGSound in~\Fref{fig:missinglabel_vis}. The examples show that our model can capture diverse contents of a video that are not annotated.

\end{document}